\documentclass{article}

    \usepackage[preprint]{neurips_2022}

\usepackage[utf8]{inputenc} % allow utf-8 input
\usepackage[T1]{fontenc}    % use 8-bit T1 fonts
\usepackage{hyperref}       % hyperlinks
\usepackage{url}            % simple URL typesetting
\usepackage{booktabs}       % professional-quality tables
\usepackage{amsfonts}       % blackboard math symbols
\usepackage{nicefrac}       % compact symbols for 1/2, etc.
\usepackage{microtype}      % microtypography
\usepackage{xcolor}         % colors
\usepackage{graphicx}
\usepackage{todonotes}
\usepackage{caption}
\usepackage{comment}
\usepackage{subcaption}
\usepackage{amsthm}
\usepackage{amsmath}
\usepackage{amssymb}
\usepackage{mathtools}
\usepackage{booktabs}            % professional-quality tables
\usepackage{multirow}            % tabular cells spanning multiple rows
\usepackage{amsfonts}            % blackboard math symbols
\usepackage{graphicx}            % figures
\usepackage{duckuments}          % sample images
\usepackage{dsfont}
\usepackage{natbib}
\usepackage{scalerel}
\usepackage[noend,ruled,vlined,noline]{algorithm2e}
\SetKwComment{tcc}{$\triangleright$~}{}
\SetCommentSty{normalfont}
\SetKwInput{KwRequire}{Require}
\usepackage{wrapfig}

\setcitestyle{numbers}
\usepackage{etoolbox}\AtBeginEnvironment{algorithm}{\tiny}

\theoremstyle{plain}
\newtheorem{theorem}{Theorem}[section]

\theoremstyle{definition}
\newtheorem{definition}[theorem]{Definition}
\newtheorem{example}[theorem]{Example}

\theoremstyle{remark}

\newcommand{\fg}[1]{\textcolor{blue}{F: #1}}

\title{Interpretable Graph Networks\\Formulate Universal Algebra Conjectures}

\author{%
Francesco Giannini$^*$\\
  Universit\`a di Siena, Italy\\
  \texttt{francesco.giannini@unisi.it} \\
  \And
  Stefano Fioravanti$^*$\\
  Universit\`a di Siena, Italy\\
  \texttt{stefano.fioravanti@unisi.it} \\
  \And
   Oguzhan Keskin\\
   University of Cambridge, UK\\
  \texttt{ok313@cam.ac.uk} \\
  \And
   Alisia Maria Lupidi\\
   University of Cambridge, UK\\
  \texttt{aml201@cam.ac.uk} \\
  \And
  Lucie Charlotte Magister \\
  University of Cambridge, UK\\
  \texttt{lcm67@cam.ac.uk} \\
  \And
  Pietro Li\'o \\
  University of Cambridge, UK\\
  \texttt{pl219@cam.ac.uk} \\
  \And
  Pietro Barbiero$^*$ \\
  University of Cambridge, UK\\
  \texttt{pb737@cam.ac.uk} \\}

\begin{document}

\maketitle

\begin{abstract}
The rise of Artificial Intelligence (AI) recently empowered researchers to investigate hard mathematical problems which eluded traditional approaches for decades. Yet, the use of AI in Universal Algebra (UA)---one of the fields laying the foundations of modern mathematics---is still completely unexplored. This work proposes the first use of AI to investigate UA's conjectures with an equivalent equational and topological characterization.
While topological representations would enable the analysis of such properties using graph neural networks, the limited transparency and brittle explainability of these models hinder their straightforward use to empirically validate existing conjectures or to formulate new ones.
To bridge these gaps, we propose a general algorithm generating AI-ready datasets based on UA's conjectures, and introduce a novel neural layer to build fully interpretable graph networks. The results of our experiments demonstrate that interpretable graph networks: (i) enhance interpretability without sacrificing task accuracy, (ii) strongly generalize when predicting universal algebra's properties, (iii) generate simple explanations that empirically validate existing conjectures, and (iv) identify subgraphs suggesting the formulation of novel conjectures. 
% Our findings demonstrate the potential of our methodology and open the doors of universal algebra to AI.

% Several problems in UA can be formalized in terms of topological properties of algebraic structures, and  are therefore suitable to be investigated by graph neural architectures. 
% In this regard, this paper proposes an approach that uses a newly devised interpretable GNN to relate equational properties of special algebraic structures to sub-portions of their graph representation. 
% %
%  Interpretable GNNs have the potential to offer explainability and transparency in the decision-making process of AI models. This is essential in mathematical research as it provides a way to understand how AI is aiding in the discovery of new theorems.
% such as the distributive and modular properties of algebraic lattices. 
% The architecture presented in this paper is capable of identifying known sub-patterns that characterize these properties while also suggesting new theorems for mathematicians to investigate. The use of Interpretable GNNs provides insight into how these suggestions are made and can aid in the development of new mathematical knowledge.
% Overall, this paper highlights the potential of AI for Mathematics, specifically in Universal Algebra, and emphasizes the importance of Interpretable GNNs in ensuring transparency and interpretability in AI-assisted mathematical research.
\end{abstract}

\begin{wrapfigure}[11]{r}{0.38\textwidth}
\vspace{-2.2cm}
    \includegraphics[width=.38\textwidth]{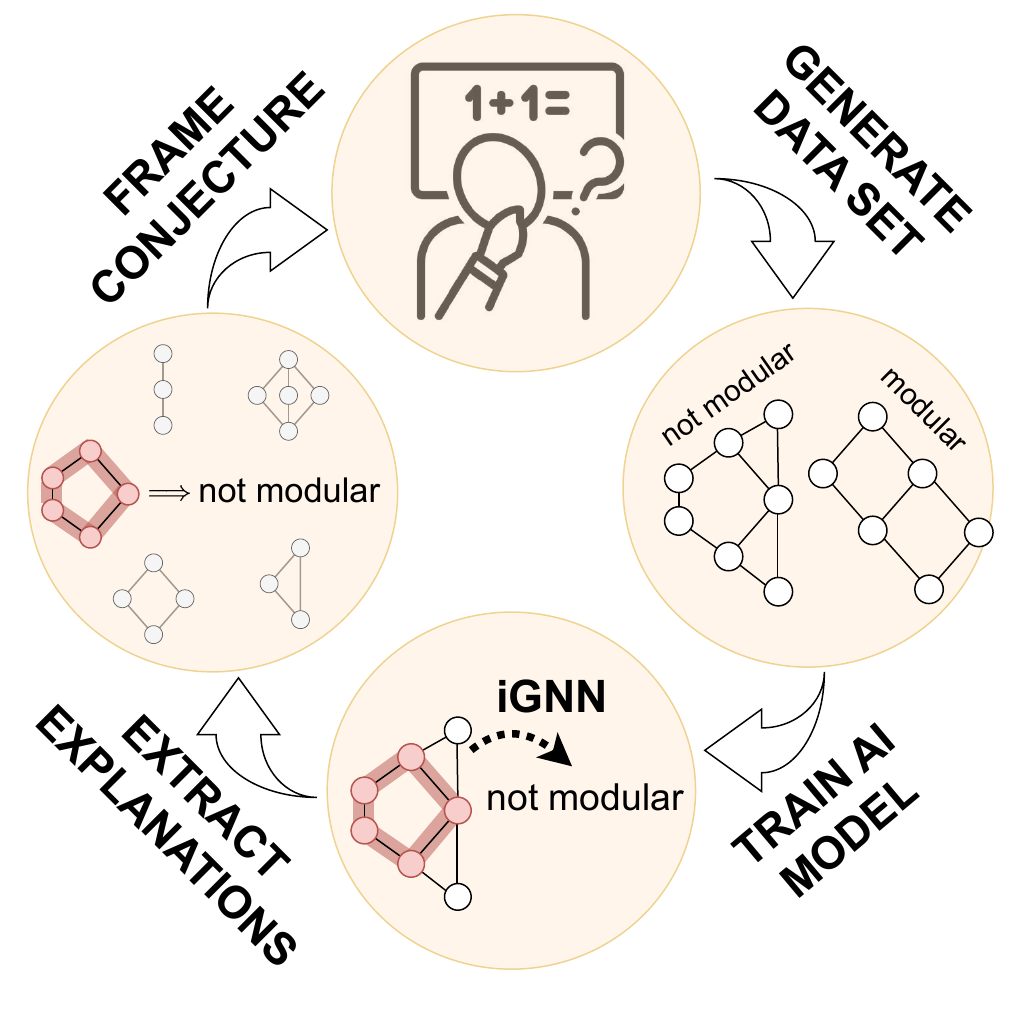}
    \caption{{\small Interpretable graph networks support universal algebra research.}}
    \label{fig:vis_abs}
\end{wrapfigure}
\section{Introduction}
Universal Algebra (UA, \citep{BurrisSanka}) is one of the foundational fields of modern mathematics, yet the complexity of studying abstract algebraic structures hinders scientific progress and discourages many academics. Recently, the emergence of powerful AI technologies empowered researchers to investigate intricate mathematical problems which eluded traditional approaches for decades, leading to the solution of open problems (e.g., ~\cite{lample2019deep}) and discovery of new conjectures (e.g., ~\cite{Davies2021}). Yet, universal algebra currently remains an uninvestigated realm for AI, a completely uncharted territory with deep impact in all mathematical disciplines.

Universal algebra studies algebraic structures from an abstract perspective. Interestingly, several UA conjectures equivalently characterize algebraic properties using equations or graphs~\citep{JipsenRose}. In theory, studying UA properties as graphs would enable the use of powerful AI techniques, such as Graph Neural Networks (GNN,~\citep{scarselli2008graph}), which excel on graph-structured data. However, two factors currently limit scientific progress. First, the \textit{absence of benchmark datasets} suitable for machine learning prevents widespread application of AI to UA. Second, \textit{GNNs' opaque reasoning} obstructs human understanding of their decision process~\citep{rudin2019stop}. Compounding the issue of GNNs' limited transparency, GNN explainability methods mostly rely on brittle and untrustworthy local/post-hoc methods \citep{huang2022graphlime,luo2020parameterized,magister2021gcexplainer,rudin2019stop,ying2019gnnexplainer} or pre-defined subgraphs for explanations~\citep{azzolin2022global, vu2020pgm}, which are often unknown in UA. 
% Hence, the limited transparency and brittle explainability of GNNs hinder universal algebrists from using them to empirically validate existing conjectures or to formulate new ones.

\textbf{Contributions.}
In this work, we investigate universal algebra's conjectures through AI (Figure~\ref{fig:vis_abs}), venturing for the first time in this previously uncharted territory. Our work includes three significant contributions. First, we propose a novel algorithm that generates a dataset suitable for training AI models based on an UA equational conjecture. Second, we generate and release the first-ever universal algebra's dataset compatible with AI, which contains more than $29,000$ lattices and the labels of $5$ key properties i.e., modularity, distributivity, semi-distributivity, join semi-distributivity, and meet semi-distributivity. And third, we introduce a novel neural layer that makes GNNs fully interpretable, according to Rudin's \cite{rudin2019stop} notion of interpretability. The results of our experiments demonstrate that interpretable GNNs (iGNNs): (i) enhance GNN interpretability without sacrificing task accuracy, (ii) strongly generalize when trained to predict universal algebra's properties, (iii) generate simple concept-based explanations that empirically validate existing conjectures, and (iv) identify subgraphs which could be relevant for the formulation of novel conjectures. Our findings demonstrate the potential of our methodology and open the doors of universal algebra to AI.

\section{Background}
% and Problem Definition}
\label{sec:back}

% \subsection{Universal Algebra}
% \label{sec:ua}
% \todo[@Stefano, @Francesco]

Universal Algebra is a branch of mathematics studying general and abstract algebraic structures. \emph{Algebraic structures} are typically represented as ordered pairs $\mathbf{A} = (A, F)$, consisting of a non-empty set $A$ and a collection of operations $F$ defined on the set. %The operations may include binary operations (such as addition or multiplication), unary operations (such as negation or inverse), nullary operations (constants), and so forth. 
UA aims to identify algebraic properties (often in equational form) shared by various mathematical systems. In particular, \emph{varieties} are classes of algebraic structures sharing a common set of identities, which enable the study of algebraic systems based on their common properties. %They form the foundation for exploring the fundamental principles and interrelationships within the field of universal algebra. 
Prominent instances of varieties that have been extensively studied across various academic fields encompass Groups, Rings, Boolean Algebras, Fields, and many others. 
% We refer the reader to \cite{BurrisSanka} for foundational notions of UA.%universal algebra has applications in various areas of mathematics and beyond. It provides a foundation for understanding the structure and properties of algebraic systems in fields like computer science, physics, and engineering. It also serves as a fundamental tool in the study of mathematical logic, model theory, and formal languages.
%\subsection{Lattices} 
%\label{sec:latt}
A particularly relevant variety of algebras are Lattices (details in Appendix \ref{def:lattice}), which are often studied for their connection with logical structures.
\begin{definition} \label{def:back_lattice}
 A \emph{lattice} $\mathbf{L}$ is an algebraic structure composed by a non-empty set $L$ and two binary operations $\vee$ and $\wedge$, satisfying the commutativity, associativity, idempotency, and absorption axioms. %following axioms and their duals obtained exchanging $\vee$ and $\wedge$:
 \begin{comment}
\begin{align*}
&x \vee y \approx y \vee x  &&\text{(commutativity)}
\\&x \vee (y \vee z) \approx (x \vee y)  &&\text{(associativity)}
\\&x \vee x \approx x  &&\text{(idempotency)}
\\&x \approx x \vee (x \wedge y)  &&\text{(absorption)}
\end{align*}
\end{comment}
\end{definition}
%class of lattice class of lattices is studied both for is connection with logical structures and for its algebraic properties. 
\begin{wrapfigure}[8]{r}{0.25\textwidth}
    \includegraphics[width=.12\textwidth,trim={1.7cm 0 0 0},clip]{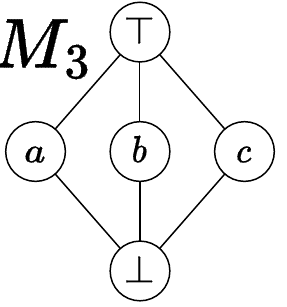}
    \includegraphics[width=.12\textwidth,trim={0 0 1.7cm 0},clip]{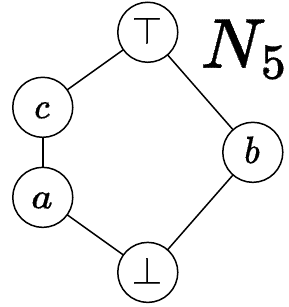}
    \caption{\small Hasse diagrams.}
    \label{fig:small_n5m3}
\end{wrapfigure}
Equivalently a lattice can be characterized as a partially ordered set in which every pair of elements has a \emph{supremum} and an \emph{infimum} (cf. Appendix \ref{def:lattice}). Lattices also have formal representations as graphs via \emph{Hasse diagrams} $(L,E)$ (e.g., Figure~\ref{fig:small_n5m3}), where each node $x \in L$ is a lattice element, and directed\footnote{The orientation of Hasse diagrams is always to be meant bottom-up, hence we will omit  arrows for simplicity.} edges $(x,y) \in E \subseteq L \times L$ represent the  ordering relation, such that if $(x,y) \in E$ then $x \leq_L y$ in the ordering of the lattice. A \emph{sublattice} $\mathbf{L}'$ of a lattice $\mathbf{L}$ is a lattice such that $L' \subseteq L$ and $\mathbf{L}'$ preserves the original order (the ``essential structure'') of $L$, i.e. for all $x, y \in L'$ then $x \leq_{L'} y$ if and only if $x \leq_L y$. The foundational work by \citet{birkhoff1935structure}, \citet{Dedekind1900}, and \citet{jonsson} played a significant role in discovering that some significant varieties of lattices can be characterized through the omission of one or more lattices. Specifically, a variety $\mathcal{V}$ of lattices is said to \emph{omit} a lattice $\mathbf{L}$ if it cannot be identified as a sublattice of any lattice in $\mathcal{V}$. A parallel line of work in UA characterizes lattices in terms of equational ("$\textit{term}_1\approx\textit{term}_2$") and quasi-equational ("if $\textit{equation}_1$ holds then $\textit{equation}_2$ holds") properties, such as distributivity and  modularity.
% This exclusion criterion serves as a useful means of characterizing and classifying different types of lattice varieties.
\begin{definition}
    Let $\mathbf{L}$ be a lattice. $\mathbf{L}$ is \emph{modular} if it satisfies $x \leq y \rightarrow x \vee (y \wedge z) \approx y \wedge (x \vee z)$; \emph{distributive} if it satisfies $x \vee (y \wedge z) \approx (x \vee y) \wedge (x \vee z)$.
    % \begin{align*}
    %     &x \leq y \rightarrow x \vee (y \wedge z) \approx y \wedge (x \vee z) &&\text{(modularity)}
    %     \\&x \vee (y \wedge z) \approx (x \vee y) \wedge (x \vee z) &&\text{(distributivity)}
    % \end{align*}
\end{definition}
For instance, as showed in Figure~\ref{fig:small_n5m3}, $\mathbf{N}_5$ is neither modular nor distributive- considering the substitution $x = a, y= c, z= b$. The same substitution shows that $\mathbf{M}_3$ is not distributive. The classes of distributive and modular lattices show classical examples of varieties that can equivalently be characterized using equations and lattice omissions, as illustrated by the following theorems. 
\begin{theorem}[\citet{Dedekind1900}]\label{the:n5}
A lattice variety $\mathcal{V}$ is modular if and only if $\mathcal{V}$ omits $\mathbf{N}_5$.
\end{theorem}
\begin{theorem}[\citet{birkhoff1935structure}]\label{the:m3}
A lattice variety $\mathcal{V}$ is distributive if and only if $\mathcal{V}$ omits $\mathbf{N}_5$ and $\mathbf{M}_3$.
\end{theorem}
Starting from these classic results, the investigation of lattice omissions and the structural characterizations of classes of lattices has evolved into a rich and extensively studied field~\citep{JipsenRose}, but it was never approached with advanced AI methods before. 

\section{Methods}
\label{sec:methods}
The problem of characterizing lattice varieties through lattice omission is very challenging as it requires the analysis of large (potentially infinite) lattices~\citep{birkhoff1935structure,Dedekind1900,jonsson}. % on small latticesand it quickly becomes intractable for large lattices
To address this task, we propose the first AI-assisted framework supporting mathematicians in finding empirical evidences to validate existing conjectures and to suggest novel theorems. To this end, we propose a general algorithm (Section \ref{sec:latt_data}) allowing researchers in universal algebra to define a property of interest and generate a dataset suitable to train AI models. We then introduce interpretable graph networks (Section \ref{sec:IGN}) which can suggest candidate lattices whose omission is responsible for the satisfaction of the given algebraic property.

\subsection{A Tool to Generate Datasets of Lattice Varieties}
\label{sec:latt_data}
% \begin{minipage}{.46\textwidth}
\begin{wrapfigure}[11]{R}{0.5\textwidth}
\vspace{-0.5cm}
\begin{algorithm}[H]
\caption{\small Generate dataset of lattice varieties.}
\label{alg:dataset}
% \SetCustomAlgoRuledWidth{0.45\textwidth}  
% \begin{algorithmic}
    % \Require
\DontPrintSemicolon
\SetInd{0.1em}{1em}
\KwIn{$n \geq 1$, \emph{hasProperty}($\cdot$,$\cdot$,$\cdot$)\tcp*[f]{\textit{$n$: cardinality}}}
$\textit{Dataset} = []$\;
$\textit{AllFuncs}\gets \textit{genAllFuncs}(n)$
\tcp*[f]{\textit{binary functions as $n\times n$ matrices}}\;
\For(\tcp*[f]{\textit{$L(i,j)=1$ meaning $i\leq_L j$}}){$L\in \textit{AllFuncs}$}
{\If(\tcp*[f]{\textit{check if $\leq_L$ is refl., antisym. and trans.}}){$\textit{isPartialOrder}(L)$}{
        \If(\tcp*[f]{\textit{check if $L$ is a lattice
        % each node pair has unique $\inf$ and $\sup$
        }}){$\textit{isLattice}(L)$}{
       \For{$i,j\leq n$}{
       $\wedge_L[i,j]\gets \sup_{x\leq L} \{ x\leq_L i \mbox{ and } x\leq_L j\}$\;
$\vee_L[i,j]\gets \inf_{x\leq L}\{ i\leq_L x\mbox{ and } j\leq_L x\}$
}
\If(\tcp*[f]{\textit{check $\wedge_L$,$\vee_L$ properties}}){$\textit{hasProperty}(L,\wedge_L,\vee_L)$}
{$\textit{Dataset}.append([L,\text{True}])$}
\Else{$\textit{Dataset}.append([L,\text{False}])$}
}
}
}
% \end{algorithmic}
\end{algorithm}
% \end{minipage}
\end{wrapfigure}
We propose a general methodology to investigate any algebraic property whose validity can be verified on a finite lattice. In this work, we focus on properties that can be characterized via equations 
and quasi-equations. To train AI models, we propose a general dataset generator\footnote{The dataset generator code and the generated datasets will be made public in case of paper acceptance.} for lattice varieties (Algorithm \ref{alg:dataset}). The generator takes as input the number of nodes $n$ in the lattices  and a function to check whether a lattice satisfies a given property. We generate $2^{n\times n}$ matrices of size $n\times n$, containing all binary functions definable on $\{1,\ldots,n\}^2$, and filter only binary matrices representing partial orders\footnote{Our algorithm optimizes this step considering only reflexive and antisymmetric binary relations, and enforces transitivity with an easy fix-point calculation.}. Then, we verify that the partial ordered set $L$ is a lattice, by checking that any pair of nodes always has a unique infimum and supremum. This directly verifies that $\wedge_L$ and $\vee_L$ satisfy Definition~\ref{def:back_lattice}. Finally, we check whether the lattice satisfies the target property or not, and append it and the property label to our dataset. We remark that checking the validity of a single ternary equation on a medium-size lattice is not computationally prohibitive (i.e., it ``only'' requires checking $n^3$ identities), but the number of existing lattices increases exponentially as $n$ increases. For instance, it is known that there are at least 2,000,000 non-isomorphic lattices with $L=10$ elements~\cite{berman2000counting}. Therefore, we only sample a fixed number of lattices per cardinality starting from a certain node cardinality. While this may seem a strong bias, we notice that known and relevant lattice omissions often rely on lattices with few nodes~\cite{birkhoff1935structure, Dedekind1900}. To empirically verify that this is not a significant limitation, in our experiments we deliberately investigate the generalization capacity of GNNs when trained on small-size lattices and tested on larger ones. This way we can use GNNs to predict the satisfiability of equational properties on large graph structures without explicitly checking them. Using Algorithm~\ref{alg:dataset}, we generated the first large-scale AI-compatible datasets of lattices containing more than $29,000$ graphs and the labels of $5$ key properties of lattice (quasi-)varieties i.e., modularity, distributivity, semi-distributivity, join semi-distributivity, and meet semi-distributivity, whose definitions can be found in Appendix \ref{app:algebra}.

\subsection{Interpretable Graph Networks (iGNNs)}
\label{sec:IGN}

In this section, we design an interpretable graph network (iGNN, Figure~\ref{fig:ignn}) that satisfies the notion of "interpretability" introduced by~\citet{rudin2019stop}. According to this definition, a machine learning (ML) system is interpretable if and only if (1) its inputs are semantically meaningful, and (2) its model inference is simple for humans to understand (e.g., sparse and/or symbolic). This definition covers ML systems that take tabular datasets or sets of concepts as inputs, and (piece-wise) linear models such as logistic regression or decision trees. To achieve this goal in GNNs, we introduce an interpretable graph layer that learns semantically meaningful concepts and uses them as inputs for a simple linear classification layer. We then show how this layer can be included into existing architectures or into hierarchical iGNNs, which consist of a sequence of interpretable graph layers.

\subsubsection{Interpretable Graph Layer}
\begin{figure}
\label{fig:gnn_arch}
    \centering
    \includegraphics[width=.9\textwidth]{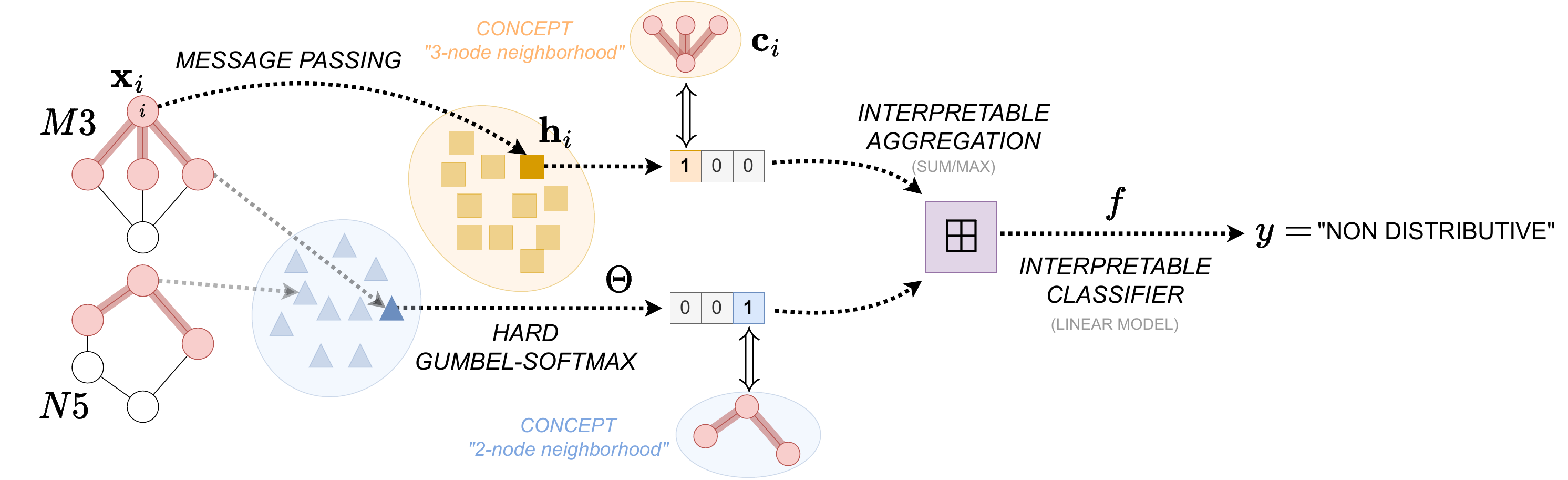}
    \caption{An interpretable graph layer (i) aggregates node features with message passing, (ii) generates a node-level concept space with a hard Gumbel-Softmax activation $\Theta$, (iii) generates a graph-level concept space with an interpretable permutation invariant pooling function $\boxplus$ on node-level concepts, and (iv) predicts a task label with an interpretable classifier $f$ using graph-level concepts.}
    \label{fig:ignn}
\end{figure}

The interpretable graph layer (Figure~\ref{fig:ignn}) serves three main functions: message passing, concept generation, and task predictions. The first step of the interpretable graph layer involves a standard message passing operation (Eq.~\ref{eq:node-concepts} right), which aggregates information from node neighbors. This operation enables to share and process relational information across nodes and it represents the basis of any GNN layer.

\paragraph{Node-level concepts.}
An interpretable concept space is the first step towards interpretability. Following~\citet{ghorbani2019towards}, a relevant concept is a ``high-level human-understandable unit of information'' shared by input samples and thus identifiable with clustering techniques. Message passing algorithms do cluster node embeddings based on the structure of node neighborhoods, as observed by~\citet{magister2021gcexplainer}. However, the real-valued large embedding representations $\mathbf{h}_i \in \mathbb{R}^q, q\in \mathbb{N}$ generated by message passing can be challenging for humans to interpret. To address this, we use a hard Gumbel-Softmax activation $\Theta: \mathbb{R}^q \mapsto \{0,1\}^q$, following~\citet{azzolin2022global}: 
\begin{equation} \label{eq:node-concepts}
\mathbf{c}_i = \Theta \big( \mathbf{h}_i \big) 
\qquad 
\mathbf{h}_i = \phi \Big(\mathbf{x}_i, \bigoplus_{j \in N_i} \psi(\mathbf{x}_i, \mathbf{x}_j)\Big)
\end{equation}
where $\psi$ and $\phi$ are learnable functions aggregating information from a node neighborhood $N_i$, and $\oplus$ is a permutation invariant aggregation function (such as sum or mean). During the forward pass, the Gumbel-Softmax activation $\Theta$ produces a one-hot encoded representation of each node embedding. Since nodes sharing the same neighborhood have similar embeddings $\mathbf{h}_i$ due to message passing, they will also have the same one-hot vector $\mathbf{c}_i$ due to the Gumbel-Softmax, and vice versa - we can then interpret nodes having the same one-hot concept $\mathbf{c}_i$ as nodes having similar embeddings $\mathbf{h}_i$ and thus sharing a similar neighborhood. 
More formally, we can assign a semantic meaning to a reference concept $\gamma \in \{0,1\}^q$ by visualizing concept prototypes corresponding to the inverse images of a node concept vector. In practice, we can consider a subset of the input lattices $\Gamma$ corresponding to the node's ($p$-hop)  neighborhood covered by message passing:
\begin{align}
    \Gamma (\gamma, p) = \Big\{ \mathbf{L}^{\langle i,p \rangle} \mid i \in L \wedge \ \mathbf{L} \in \mathcal{D} \wedge \ \mathbf{c}_i = \gamma \Big\}
\end{align}
where $\mathcal{D}$ is the set of all training lattices, and $\mathbf{L}^{\langle i,p \rangle}$ is the graph corresponding to the $p$-hop neighborhood ($p \in \{1,\dots,|L|\}$) of the node $i \in L$, as suggested by~\citet{ghorbani2019towards,magister2021gcexplainer}. This way, by visualizing concept prototypes as subgraph neighborhoods, the meaning of the concept representation becomes easily interpretable to humans (Figure~\ref{fig:ignn}), aiding in the understanding of the reasoning process of the network.

\begin{example}[Interpreting node-level concepts]
\label{ex:nc}
Consider the problem of classifying distributive lattices with a simplified dataset including $\mathbf{N}_5$ \includegraphics[scale=0.08]{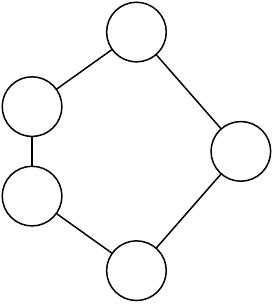} and $\mathbf{M}_3$ \includegraphics[scale=0.08]{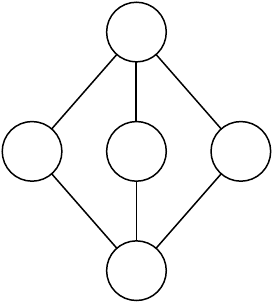} only, and where each node has a constant feature $x_i = 1$. As these two lattices only have nodes with 2 or 3 neighbours, one layer of message passing will then generate only two types of node embeddings e.g., $\mathbf{h}_{II} = [0.2, -0.4, 0.3]$ for nodes with a 2-nodes neighborhood (e.g., \includegraphics[scale=0.08]{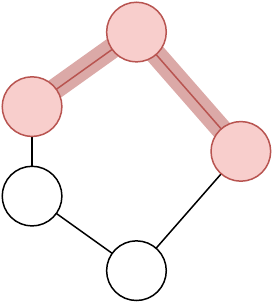}), and $\mathbf{h}_{III} = [0.6, 0.2, -0.1]$ for nodes with a 3-nodes neighborhood (e.g., \includegraphics[scale=0.08]{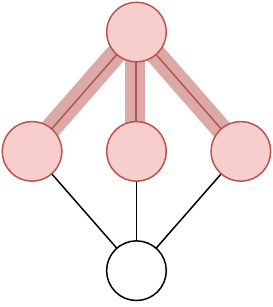}).
As a consequence, the Gumbel-Softmax will only generate two possible concept vectors e.g., $\mathbf{c}_{II} = [0, 0, 1]$ and $\mathbf{c}_{III} = [1, 0, 0]$. 
Hence, for instance the concept \includegraphics[scale=0.08]{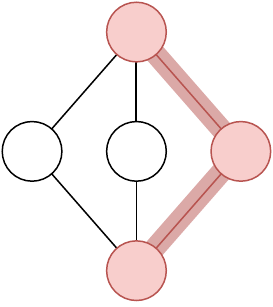} belongs to $\mathbf{c}_{II}$, while \includegraphics[scale=0.08]{fig/_base/lattices_m3_3nodes.pdf} belongs to $\mathbf{c}_{III}$.
% The one-to-one mapping between the structure of the neighborhood and the concept activation makes the learned concepts easily interpretable e.g., $\mathbf{c}_i = [1,0,0] \iff$ \includegraphics[scale=0.08]{fig/_base/lattices_m3_3nodes.pdf} or $\mathbf{c}_j = [0,0,1] \iff$ \includegraphics[scale=0.08]{fig/_base/lattices_m3_2nodes.pdf}. 
\end{example}

\paragraph{Graph-level concept embeddings.}
To generate a graph-level concept space in the interpretable graph layer, we can utilize the node-level concept space produced by the Gumbel-Softmax. Normally, graph-level embeddings are generated by applying a permutation invariant aggregation function on node embeddings. However, in iGNNs we restrict the options to (piece-wise) linear permutation invariant functions in order to follow our interpretability requirements dictated by~\citet{rudin2019stop}. This restriction still includes common options such as max or sum pooling. Max pooling can easily be interpreted by taking the component-wise max over the one-hot encoded concept vectors $\mathbf{c}_i$. After max pooling, the graph-level concept vector has a value of $1$ at the $k$-th index if and only if at least one node activates the $k$-th concept i.e., $\exists i \in L, \mathbf{c}_{ik} = 1$. Similarly, we can interpret the output of a sum pooling: a graph-level concept vector takes a value $v \in \mathbb{N}$ at the $k$-th index after sum pooling if and only if there are exactly $v$ nodes activating the $k$-th concept i.e., $\exists i_0,\dots,i_v \in L, \mathbf{c}_{ik} = 1$.

\begin{example}[Interpreting graph-level concepts]
\label{ex:gc}
Following Example \ref{ex:nc}, let us use sum pooling to generate graph-level concepts. For an $\mathbf{N}_5$ graph, we have 5 nodes with exactly the same 2-node neighborhood. Therefore, sum pooling generates a graph-level embedding $[0,0,5]$, which certifies that we have 5 nodes of the same type e.g., \includegraphics[scale=0.08]{fig/_base/lattices_n5_2nodes.pdf}. For an $\mathbf{M}_3$ graph, the top and bottom nodes have a 3-node neighborhood e.g., \includegraphics[scale=0.08]{fig/_base/lattices_m3_3nodes.pdf}, while the middle nodes have a 2-node neighborhood e.g., \includegraphics[scale=0.08]{fig/_base/lattices_m3_2nodes.pdf}. This means that sum pooling generates a graph-level embedding $[2, 0, 3]$, certifying that we have 2 nodes of type \includegraphics[scale=0.08]{fig/_base/lattices_m3_3nodes.pdf} and 3 nodes of type  \includegraphics[scale=0.08]{fig/_base/lattices_m3_2nodes.pdf}.
\end{example}

% \paragraph{Concept visualization}

\paragraph{Interpretable classifier.}
To prioritize the identification of relevant concepts, we use a classifier to predict the task labels using the concept representations. A black-box classifier like a multi-layer perceptron~\cite{mlp} would not be ideal as it could compromise the interpretability of our model, so  instead we use an interpretable linear classifier such as a single-layer network~\cite{slp_Karimboyevich_Nematullayevich_2022}. This allows for a completely interpretable and differentiable model from the input to the classification head, as the input representations of the classifier are interpretable concepts and the classifier is a simple linear model which is intrinsically interpretable as discussed by~\citet{rudin2019stop}. In fact, the weights of the perceptron can be used to identify which concepts are most relevant for the classification task. Hence, the resulting model can be used not only for classification, but also to interpret and understand the problem at hand.

\subsubsection{Interpretable architectures}
The interpretable graph layer can be used to instantiate different types of iGNNs. One approach is to plug this layer as the last message passing layer of a standard GNN architecture: 
\begin{align}
& \hat{y} = f\Big(\boxplus_{i \in K} \ \ \Big(\Theta \Big(\phi^{(K)} \Big(\mathbf{h}_i^{(K-1)}, \bigoplus_{j \in N_i} \psi^{(K)}(\mathbf{h}_i^{(K-1)}, \mathbf{h}_j^{(K-1)})\Big)\Big)\Big)\Big) \\
& \mathbf{h}_i^{(l)} = \phi^{(l)} \Big(\mathbf{h}_i^{(l-1)}, \bigoplus_{j \in N_i} \psi^{(l)}(\mathbf{h}_i^{(l-1)}, \mathbf{h}_j^{(l-1)})\Big)
\quad l = 1,\dots,K
\end{align}
where $f$ is an interpretable classifier (e.g., single-layer network), $\boxplus$ is an interpretable piece-wise linear and permutation-invariant function (such as max or sum), $\Theta$ is a Gumbel-Softmax hard activation function, and $\mathbf{h}_i^0 = \mathbf{x}_i$. In this way, we can interpret the first part of the network as a feature extractor generating well-clustered latent representations from which concepts can be extracted. This approach is useful when we only care about the most complex neighborhoods/concepts. Another approach is to generate a hierarchical transparent architecture where each GNN layer is interpretable:
\begin{equation}
\hat{y}^{(l)} = f\Big(\boxplus_{i \in K} \Big(\ \Theta \Big(\mathbf{h}_j^{(l)} \Big) \Big)\Big)
\qquad l = 1,\dots,K
\end{equation}
In this case, we can interpret every single layer of our model with concepts of increasing complexity. The concepts extracted from the first layer represent subgraphs corresponding to the $1$-hop neighborhood of a node, those extracted at the second layer will correspond to $2$-hop neighborhoods, and so on. These hierarchical iGNNs can be useful to get insights into concepts with different granularities. By analyzing the concepts extracted at each layer, we gain a better understanding of the GNN inference and of the importance of different (sub)graph structures for the classification task.

\subsubsection{Training}
The choice of the activation and loss functions iGNNs depends on the nature of the task at hand and does not affect their interpretability. For classification tasks, we use standard activation functions such as softmax or sigmoid, along with standard loss functions like cross-entropy. For hierarchical iGNNs (HiGNNs), we apply the loss function at each layer of the concept hierarchy, as their layered architecture enables intermediate supervisions. This ensures that each layer is doing its best to extract the most relevant concepts to solve the task. Internal losses can also be weighted differently to prioritize the formation of optimal concepts of a specific size, allowing the HiGNN to learn in a progressive and efficient way. 
% By doing so, we can train the HiGNN in a progressive and effective manner, allowing us to uncover concepts with different granularities, and gain insights into how the model is performing at different levels of abstraction.

% \subsection{}

\section{Experimental Analysis}
\label{sec:ex}
\subsection{Research questions}
In this section we analyze the following research questions:
\begin{itemize}
    \item \textbf{Generalization} - Can GNNs generalize when trained to predict universal algebra's properties? Can interpretable GNNs generalize as well?
    \item \textbf{Interpretability} - Do interpretable GNNs concepts empirically validate universal algebra's conjectures? How can concept-based explanations suggest novel conjectures? 
    % Can we use interpretable GNNs to provide insights on mutual relationships between lattice properties?
    % \item \textbf{Concept Quality} - Are the concepts identified by interpretable GNNs pure? Is the concept space complete?
\end{itemize}

\subsection{Setup}
\paragraph{Baselines.}
For our comparative study, we evaluate the performance of iGNNs and their hierarchical version against equivalent GNN models (i.e., having the same hyperparameters such as number layers, training epochs, and learning rate). For vanilla GNNs we resort to common practice replacing the Gumbel-Softmax with a standard leaky ReLU activation. We exclude from our main baselines prototype or concept-based GNNs pre-defining graph structures for explanations, as for most datasets these structures are unknown. Appendix~\ref{app:baselines} covers implementation details. We show more extensive results including local and post-hoc explanations in Appendix~\ref{app:local-xai}.
% Similarly, we do not consider local explainability methods such as GNNExplainer or integrated gradients, as our focus is to understand the behavior of the GNN on a global scale. 
% We also avoid comparing against post-hoc methods, such as GLGExplainer or GCExplainer~\citep{magister2021gcexplainer}, since we are interested in obtaining the exact and precise reason for the GNN's predictions, without relying on surrogate models to approximate the decision-making mechanism.

\paragraph{Evaluation.}
We employ three quantitative metrics to assess a model's generalization and interpretability. We use the Area Under the Receiver Operating Characteristic (AUC ROC) curve to assess task generalization. We evaluate generalization under two different conditions: with independently and identically distributed train/test splits, and out-of-distribution by training on graphs up to eight nodes, while testing on graphs with more than eight nodes (``strong generalization''). We further assess generalization under binary and multilabel settings (classifying 5 properties of a lattice at the same time). To evaluate interpretability, we use standard metrics such as completeness~\citep{yeh2020completeness} and fidelity~\citep{ribeiro2016should}. Completeness\footnote{We assess the recall of the completeness as the datasets are very unbalanced towards the negative label.} assesses the quality of the concept space on a global scale using an interpretable model to map concepts to tasks, while fidelity measures the difference in predictions obtained with an interpretable surrogate model and the original model. Finally, we evaluate the meaningfulness of our concept-based explanations by visualizing and comparing the generated concepts with ground truth lattices like e.g. $\mathbf{M}_3$ and $\mathbf{N}_5$, whose omission is known to be significant for modular and distributive properties. All metrics in our evaluation, across all experiments, are computed on test sets using $5$ random seeds, and reported using the mean and $95\%$ confidence interval.

\section{Key Findings}
% \subsection{Generalization}

\subsection{Generalization}
\begin{figure}[!h]
    \centering
    \includegraphics[width=\textwidth]{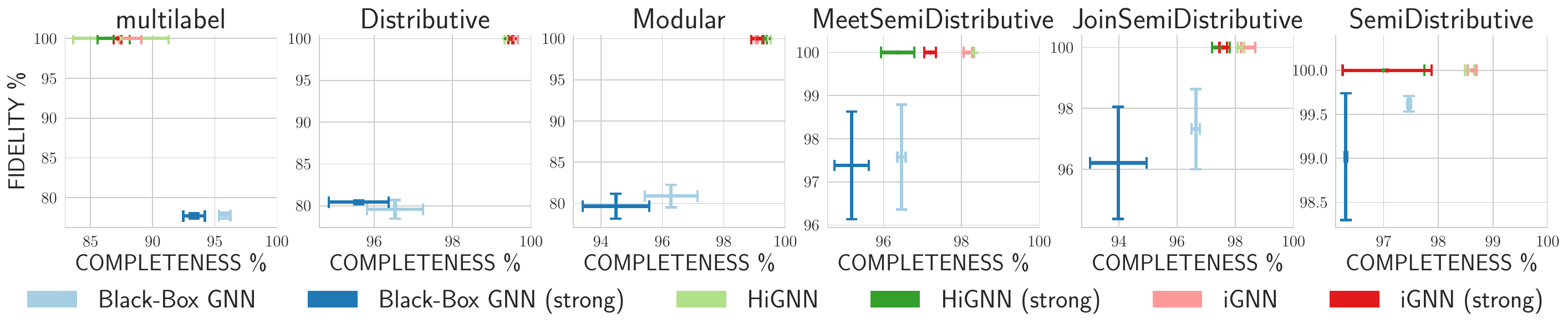}
    \caption{Accuracy-interpretability trade-off in terms of concept completeness (accuracy) and model fidelity (interpretability). iGNNs attain optimal fidelity as model inference is inherently interpretable, outmatching equivalent black-box GNNs. All models attain similar results in terms of completeness.}
    \label{fig:comp-fidelity}
\end{figure}

\paragraph{iGNNs improve interpretability without sacrificing task accuracy (Figure~\ref{fig:comp-fidelity}).}
Our experimental evaluation reveals that interpretable GNNs are able to strike a balance between completeness and fidelity, two crucial metrics that are used to assess generalization-interpretability trade-offs~\citep{ribeiro2016should}. 
% The completeness score evaluates the quality of the concept space, while the fidelity score measures the accuracy of the predictions obtained with an interpretable classifier instead of a black-box. 
We observe that the multilabel classification scenario, which requires models to learn a more varied and diverse set of concepts, is the most challenging and results in the lowest completeness scores on average. We also notice that the more challenging out-of-distribution scenario results in the lowest completeness and fidelity scores across all datasets. More importantly, our findings indicate that iGNNs achieve optimal fidelity scores, as their classification layer consists of a simple linear function of the learnt concepts which is intrinsically interpretable~\citep{rudin2019stop}.
% , aligning with the existing literature which suggests that a model with fidelity score of $100\%$ is considered to be interpretable. 
On the contrary, interpretable surrogate models of black-box GNNs exhibit, as expected, lower fidelity scores, confirming analogous observations in the explainable AI literature~\citep{ribeiro2016should,rudin2019stop}. In practice, this discrepancy between the original black-box predictions and the predictions obtained with an interpretable surrogate model questions the actual usefulness of black-boxes when interpretable alternatives achieve similar results in solving the problem at hand, as extensively discussed by~\citet{rudin2019stop}. Overall, these results demonstrate how concept spaces are highly informative to solve universal algebra's tasks and how the interpretable graph layer may improve GNNs' interpretability without sacrificing task accuracy. We refer the reader to Appendix \ref{app:completeness} for detailed discussion on quantitative analysis of concept space obtained by iGNNs under different generalization settings with comparisons to their black-box counterparts. 

\begin{wrapfigure}[14]{r}{0.7\textwidth}
    \centering
    \includegraphics[width=.7\textwidth]{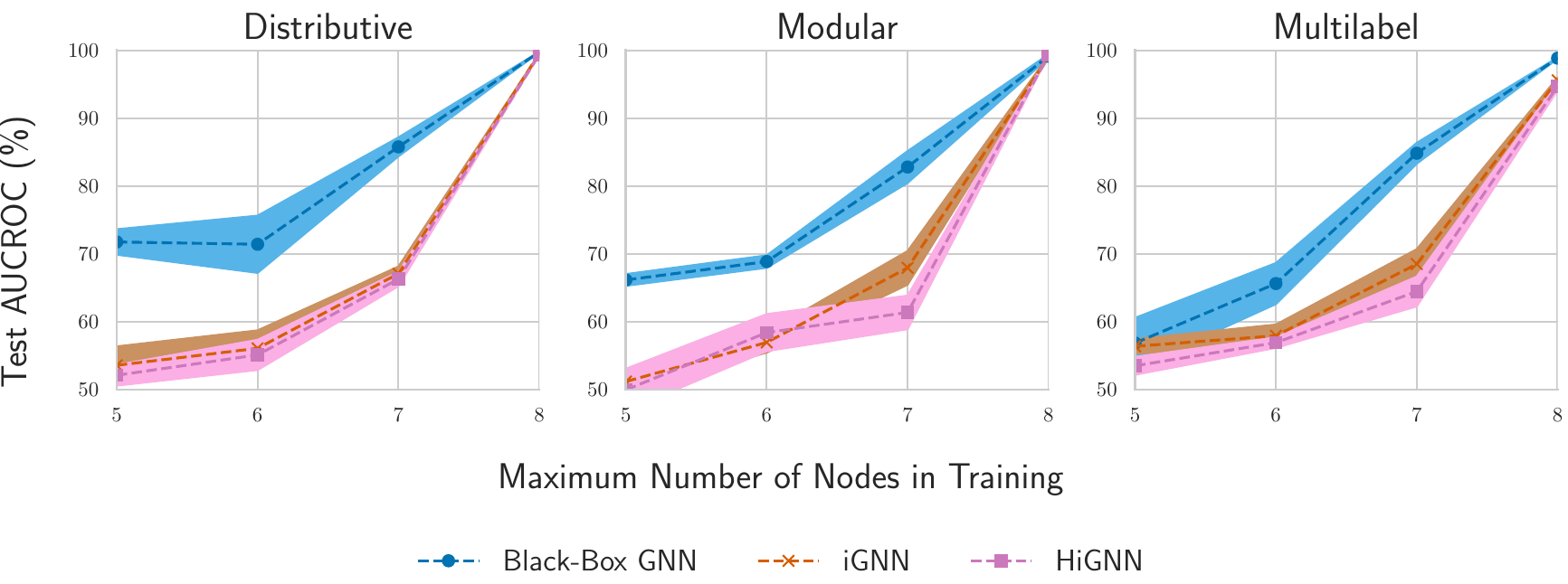}
    \caption{Strong generalization performance with respect to the maximum number of nodes used in training.}
    \label{fig:strong-generalization}
\end{wrapfigure}

\paragraph{GNNs strongly generalize on universal algebra's tasks (Figure~\ref{fig:strong-generalization}).}
Our experimental findings demonstrate the strong generalization capabilities of GNNs across the universal algebra tasks we designed. Indeed, we stress GNNs test generalization abilities by training the models on graphs of size up to $n$ (with $n$ ranging from $5$ to $8$), and evaluating their performance on much larger graphs of size up to $50$. We designed this challenging experiment in order to understand the limits and robustness of interpretable GNNs when facing a significant data distribution shift from training to test data. 
Remarkably, iGNNs exhibit robust generalization abilities (similar to their black-box counterparts) when trained on graphs up to size $8$ and tested on larger graphs. This evidence confirms the hypothesis that interpretable models can deliver reliable and interpretable predictions, as suggested by \citet{rudin2019stop}. However, we observe that black-box GNNs slightly outperform iGNNs when trained on even smaller lattices.
We hypothesize that this is due to the more constrained architecture of iGNNs, which imposes tighter bounds on their expressiveness when compared to standard black-box GNNs.
Notably, training with graphs of size up to $5$ or $6$ significantly diminishes GNNs generalization in the tasks we designed. We hypothesize that this is due to the scarcity of non-distributive and non-modular lattices during training, but it may also suggest that some patterns of size $7$ and $8$ might be quite relevant to generalize to larger graphs. Unfortunately, running generalization experiments with $n \leq 4$ was not possible since all such lattices trivially omitted $\mathbf{N}_5$ and $\mathbf{M}_3$. It is worth mentioning that GNNs performed well even in the challenging multilabel case, where they had to learn a wider and more diverse set of concepts and tasks. In all experiments, we observe a plateau of the AUC ROC scores for $n=8$, thus suggesting that a training set including graphs of this size might be sufficient to learn the relevant patterns allowing the generalization to larger lattice varieties. For detailed numerical results across all tasks, we refer the reader to Table~\ref{tab:auc} in Appendix~\ref{app:generalization}. Overall, these results emphasize the potential of GNNs in addressing complex problems in universal algebra, providing an effective tool to handle lattices that are difficult to analyze manually with pen and paper.

\subsection{Interpretability}
\begin{wrapfigure}[14]{r}{0.65\textwidth}
    \centering
    \vspace{-4mm}
    \includegraphics[width=.32\textwidth]{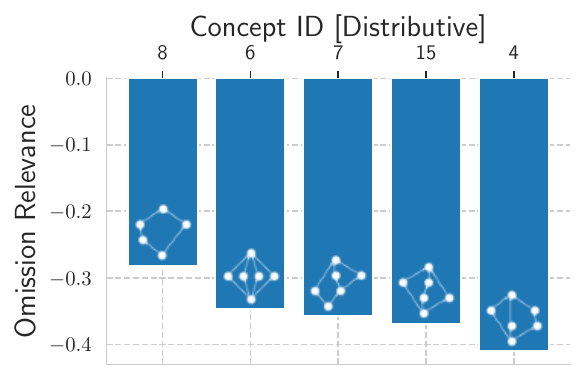}
    \includegraphics[width=.32\textwidth]{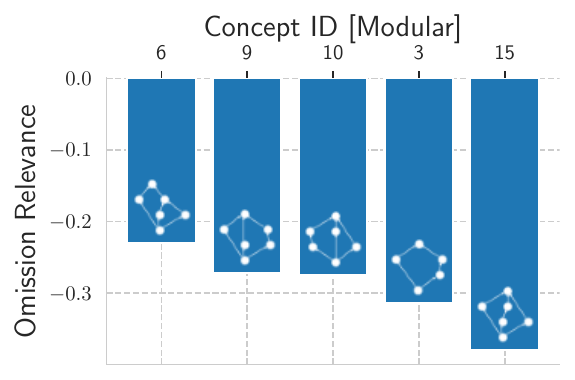}
    \caption{Ranking of relevant clusters of lattices (x-axis) according to the interpretable GNN linear classifier weights (y-axis, the lower the more relevant the cluster). $\mathbf{N}_5$ is always the most important lattice to omit for modularity, while both $\mathbf{M}_3$ and $\mathbf{N}_5$ are relevant for distributivity, thus validating theorems~\ref{the:n5} and~\ref{the:m3}.}
    \label{fig:concept-ranking}
\end{wrapfigure}

\paragraph{Concept-based explanations empirically validate universal algebra's conjectures (Figure~\ref{fig:concept-ranking}).}
We present empirical evidence to support the validity of theorems \ref{the:n5} and \ref{the:m3} by examining the concepts generated for modular and distributive tasks. For this investigation we leverage the interpretable structure of iGNNs. Similarly to~\citet{ribeiro2016should}, we visualize in Figure~\ref{fig:concept-ranking} the weights of our trained linear classifier representing the relevance of each concept. We remark that the visualization is limited to the (top-$5$) most negative weights, as we are interested in those concepts that negatively affect the prediction of a property. In the same plot, we also show the prototype of each concept represented by the 2-hop neighborhood of a node activating the concept, following a similar procedure as ~\citet{ghorbani2019towards,magister2021gcexplainer,azzolin2022global}. Using this visualization, we investigate the presence of certain concepts when classifying modular and distributive lattices. 
For the modularity task, our results show that the lattice $\mathbf{N}_5$ appears among non-modular concepts, but is never found in modular lattices, while the lattice $\mathbf{M}_3$ appears among both modular and non-modular concepts, which is consistent with Theorem \ref{the:n5}. In the case of distributivity, we observe that both $\mathbf{M}_3$ and $\mathbf{N}_5$ are present among non-distributive concepts, and are never found in distributive lattices, which is also in line with Theorem \ref{the:m3}.
These findings provide a large-scale empirical evidence for the validity of theorems \ref{the:n5} and \ref{the:m3}, and further demonstrate the effectiveness of graph neural networks in learning and analyzing lattice properties. Overall, these results highlight how interpretable GNNs can not only learn the properties of universal algebra but also identify structures that are unique to one type of lattice (e.g., non-modular) and absent from another (e.g., modular), thus providing human-interpretable explanations for what the models learn.

\begin{wrapfigure}[8]{r}{0.65\textwidth}
    \centering
    \vspace{-3.5mm}
    \includegraphics[width=.32\textwidth]{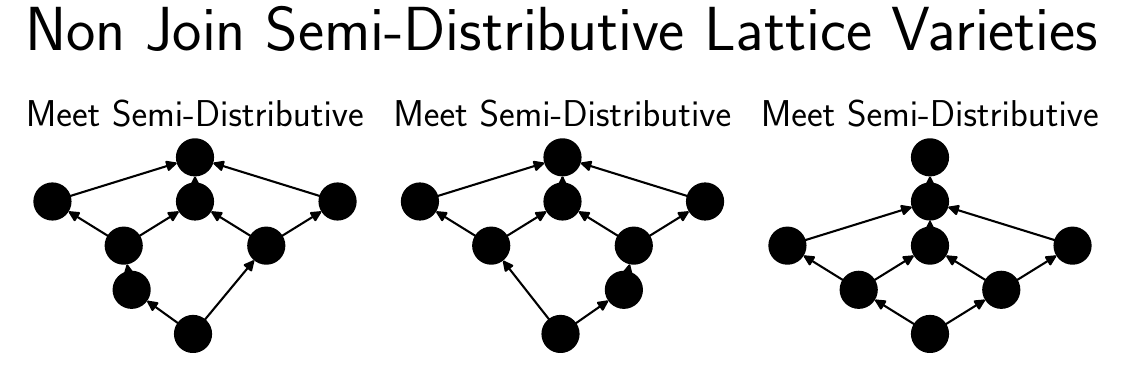}
    \includegraphics[width=.32\textwidth]{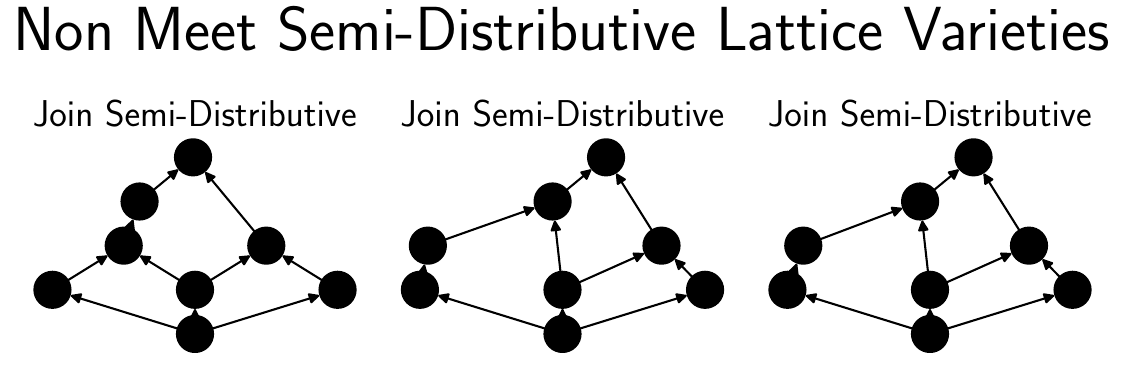}
    \caption{Contrastive explanations showing lattice varieties with a a pair of discording labels to highlight the key difference between join and meet semi-distributivity.}
    \label{fig:contrastive}
\end{wrapfigure}

\paragraph{Contrastive explanations highlight topological differences between properties of lattice varieties (Figure~\ref{fig:contrastive}).}
% With evidence that our method works (as shown for modular and distributive tasks), 
We leverage interpretable GNNs to analyze the key topological differences of classical lattice properties such as join and meet semi-distribuitivity characterized by relevant quasi-equations (cf. Appendix~\ref{def:semis}).  %gain insights into other universal algebra tasks that currently lack conjectures. %For example, the meet-semi-distributive property currently lacks a conjecture identifying the (set of) lattice(s) whose omission makes a lattice meet-semi-distributive. 
To this end, we visualize specific concept prototypes corresponding to lattices that are not meet semi-distributive against lattices that are meet semi-distributive.
We observe $\mathbf{N}_5$ but not $\mathbf{M}_3$ among the concepts of meet semi-distributive lattices, while we observe both $\mathbf{N}_5$ and $\mathbf{M}_3$ only in concepts that are not meet semi-distributive. This observation suggests that $\mathbf{N}_5$ is not a key lattice for meet semi-distributive lattices, \emph{unlike distributive lattices}. Furthermore, we find that the lattice pattern \includegraphics[width=0.03\textwidth]{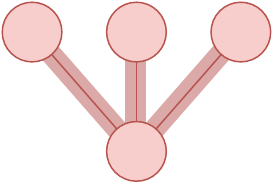} is relevant for non meet semidistributivity, while its dual \includegraphics[width=0.03\textwidth]{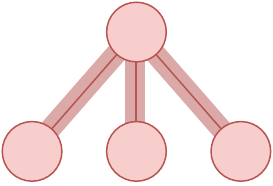} is relevant for non join semidistributivity, thus empirically confirming the hypotheses of \citet{JonssonRival}. These findings are significant because they demonstrate how analyzing concepts in interpretable GNNs can provide universal algebraists with a powerful and automatic tool to formulate new conjectures based on identifying simple lattices that play a role in specific properties. By leveraging the power of interpretable GNNs, we may uncover previously unknown connections between different properties and identify new patterns and structures that could lead to the development of new conjectures and theorems in universal algebra, providing exciting opportunities for future research in universal algebra.

\section{Discussion}

\paragraph{Relations with Graph Neural Network explainability.}
Graph Neural Networks (GNNs,\citep{scarselli2008graph}) process relational data generating node representations by combining the information of a node with that of its neighbors, thanks to a general learning paradigm known as message passing~\cite{gilmer2017neural}. A number of post-hoc explainability techniques have been proposed to explain the reasoning of GNNs. Inspired by vision approaches~\citep{simonyan2013deep,ribeiro2016should,ghorbani2019towards}, early explainability techniques focused on feature importance~\citep{pope2019explainability}, while subsequent works aimed to extract local explanations~\citep{ying2019gnnexplainer,luo2020parameterized,vu2020pgm} or global explanations using conceptual subgraphs by clustering the activation space~\citep{magister2021gcexplainer,zhang2022protgnn,magister2022encoding}. However, all these techniques either rely on pre-defined subgraphs for explanations (which are often unknown in UA) or provide post-hoc explanations which may be brittle and unfaithful as extensively demonstrated by~\citet{rudin2019stop}. On the contrary, our experiments show that iGNNs generate interpretable predictions according to~\citet{rudin2019stop} notion of interpretability via linear classifiers applied on sparse human-understandable concept representations.

\paragraph{Limitations.}

The approach proposed in this paper focuses on  universal algebra conjectures characterized both algebraically and topologically. Our methodology is limited to finite lattices, which may not capture all relevant information about infinite algebraic structures. However, the insights gained from finite-lattice explanations can still provide valuable information regarding a given problem (albeit with potentially limited generalization).
Moreover, our approach is restricted to topological properties on graphs, while non-structural properties 
% not captured by our approach 
may require the adoption of other kinds of (interpretable) models. 
% To address this limitation, we plan to explore different types of explainers in future research. By doing so, we aim to develop a more comprehensive understanding of the properties and structures of lattices and other algebraic structures.

% This study presents a novel approach that employs artificial intelligence (AI) to investigate conjectures in universal algebra, which are characterized both algebraically and topologically. 
% However, our methodology is limited to finite structures, which may not capture all relevant information about infinite structures. Nevertheless, the insights gained from our finite-lattice explanations can provide valuable information regarding a given problem, albeit with potentially limited generalization.
% %
% Moreover, our approach is restricted to structural properties, and non-structural properties may not be captured by our explainer. To address this limitation, we plan to explore different types of explainers in future research. By doing so, we aim to develop a more comprehensive understanding of the properties and structures of lattices and other algebraic structures.

\paragraph{Broader impact and perspectives.}
AI techniques are becoming increasingly popular for solving previously intractable mathematical problems and proposing new conjectures~\cite{lampledeep, loveland2016automated,Davies2021,he2022machine}. However, the use of modern AI methods in universal algebra was a novel and unexplored field until the development of the approach presented in this paper. To this end, our method uses interpretable graph networks to suggest graph structures that characterize relevant algebraic properties of lattices. With our approach, we empirically validated~\citet{Dedekind1900} and~\citet{birkhoff1935structure} theorems on distributive and modular lattices, by recovering relevant lattices. This approach can be readily extended---beyond equational properties determined by the omission of a sublattice in a variety~\cite{whitman1941free}---to any structural property of lattices, including the characterization of congruence lattices of algebraic varieties~\cite{AgBaFi,KearnesKiss2013,Nation1974,whitman1941free}. Our methodology can also be applied (beyond universal algebra) to investigate (almost) any mathematical property that can be topologically characterized on a graph, such as the classes of graphs/diagraphs with a fixed set of polymorphisms~\cite{Barto1,Barto,Olsak}. However, as universal algebra is a foundational branch of modern mathematics, any contribution to this field can already have significant implications in various mathematical disciplines.

\paragraph{Conclusion.}

This paper presents the first-ever AI-assisted approach to investigate equational and topological conjectures in the field of universal algebra. To this end, we present a novel algorithm to generate datasets suitable for AI models to study equational properties of lattice varieties. While topological representations would enable the use of graph neural networks, the limited transparency and brittle explainability of these models hinder their use in validating existing conjectures or proposing new ones. For this reason, we introduce a novel neural layer to build fully interpretable graph networks to analyze the generated datasets. The results of our experiments demonstrate that interpretable graph networks: enhance interpretability without sacrificing task accuracy, strongly generalize when predicting universal algebra's properties, generate simple explanations that empirically validate existing conjectures, and identify subgraphs suggesting the formulation of novel conjectures. These promising results demonstrate the potential of our methodology, opening the doors of universal algebra to AI with far-reaching impact across all mathematical disciplines.

%\section*{References}

% References follow the acknowledgments. Use unnumbered first-level heading for
% the references. Any choice of citation style is acceptable as long as you are
% consistent. It is permissible to reduce the font size to \verb+small+ (9 point)
% when listing the references.
% Note that the Reference section does not count towards the page limit.
% \medskip

\bibliographystyle{plainnat}
\bibliography{references}

\newpage

\appendix

\section{Algebra definitions}
\label{app:algebra}

\subsection{Formal defintions for Universal Algebra}
Universal algebra is the field of mathematics that studies algebraic structures, which are defined as a set $A$ along with its own collection of operations. An $n$-ary operation on $A$ is a function that takes $n$ elements of $A$ and returns a single element from the set. More formally \cite{BurrisSanka,day_1969,Jonnson_1967}:\\
\begin{definition}\textbf{N-ary function}
For $A$ non-empty set and $n$ nonnegative integer we define $A^0 = \{\emptyset\}$ and, for $n > 0$, $A^n$
is the set of n-tuples of elements from $A$. An $n$-ary operation (or function)
on $A$ is any function $f$ from $A^n$
to $A$; $n$ is the arity (or rank) of $f$. An operation $f$ on $A$ is called an n-ary operation if its arity  is $n$.\\
\end{definition}
\begin{definition}\textbf{Algebraic Structure} An algebra $ \mathcal{A}$ is a pair $ \langle A, F \rangle$ where $A$ is a non-empty set called universe and $F$ is a set of finitary operations on $A$.\\
\end{definition}

Apart from the operations on $A$, an algebra is further defined by axioms, that in the particular case of universal algebras are often of the form of identities. The collection of algebraic structures defined by equational laws are called varieties. \cite{HYLAND2007437}\\
\begin{definition}\textbf{Variety}
A nonempty class K of algebras of type $\mathcal{F}$ is called a variety if it is closed under subalgebras, homomorphic images, and direct products.
\end{definition}

\begin{definition}
\label{def:lattice}
 A \emph{lattice} $\mathbf{L}$ is an algebraic structure composed by a non-empty set $L$ and two binary operations $\vee$ and $\wedge$ satisfying the following axioms and their duals obtained exchanging $\vee$ and $\wedge$:
\begin{align*}
&x \vee y \approx y \vee x  &&\text{(commutativity)}
\\&x \vee (y \vee z) \approx (x \vee y)  &&\text{(associativity)}
\\&x \vee x \approx x  &&\text{(idempotency)}
\\&x \approx x \vee (x \wedge y)  &&\text{(absorption)}
\end{align*}
\end{definition}

\begin{theorem}[\cite{BurrisSanka}]
\label{th:pos-latt}
A partially ordered set $L$ is a lattice if and only if for every $a, b \in L$ both \emph{supremum} and \emph{infimum} of $\{a,b\}$ exist
(in $L$) with $a \vee b$ being the supremum and $a \wedge b$ the infimum.
\end{theorem}

\begin{definition}
\label{def:semis}
    Let $\mathbf{L}$ be a lattice. Then $\mathbf{L}$ is \emph{modular}(\emph{distributive}, $\vee$-\emph{semi-distributive}, $\wedge$-\emph{semi-distributive}) if it satisfies the following equation: 
     \begin{align*}
         &x \leq y \rightarrow x \vee (y \wedge z) \approx y \wedge (x \vee z) &&\text{(modularity)}
         \\&x \vee (y \wedge z) \approx (x \vee y) \wedge (x \vee z) &&\text{(distributivity)}
         \\&x \vee y \approx x \vee z \rightarrow  x \vee (y \wedge z) \approx x \vee y &&\text{($\vee$-semi-distributivity)}
         \\&x \wedge y \approx x \wedge z \rightarrow  x \wedge (y \vee z) \approx x \wedge y  &&\text{($\wedge$-semi-distributivity)}.
     \end{align*}
     Furthermore a lattice $\mathbf{L}$ is semi-distributive if is both $\vee$-semi-distributive and $\wedge$-semi-distributive
\end{definition}

\begin{figure}[!ht]
\vskip 0.2in
\begin{center}
\includegraphics[width=0.7\columnwidth]{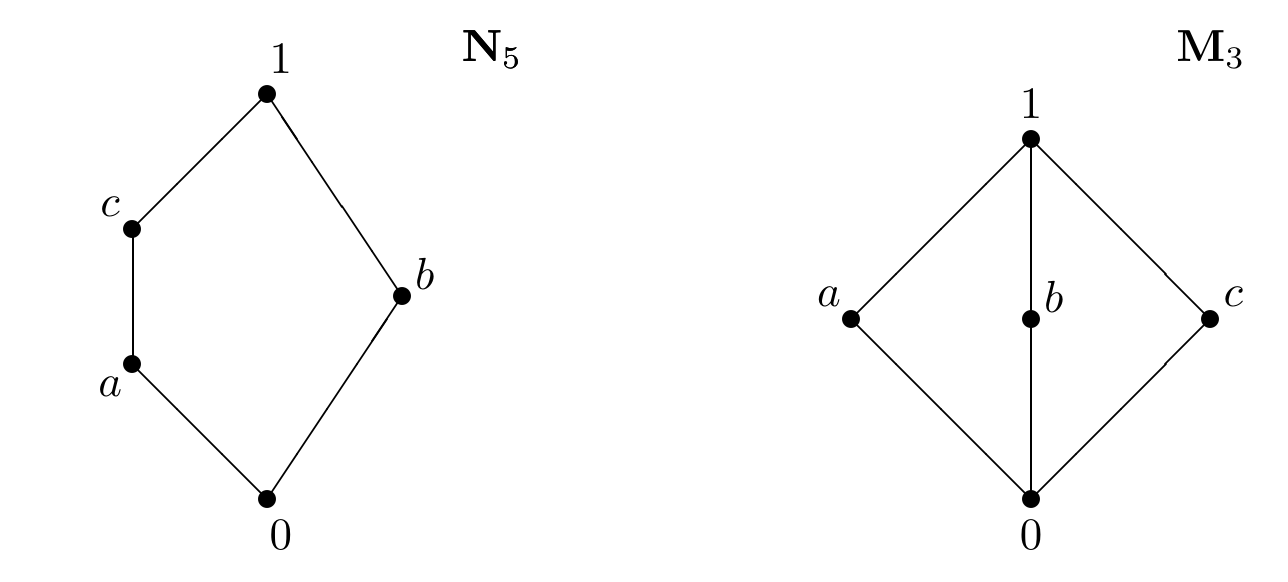}
\caption{$\mathbf{N}_5$, a non-modular non-distributive and $\mathbf{M}_3$, a modular non-distributive lattice.}
\label{fig:n5m3}
\end{center}
\vskip -0.2in
\end{figure}

Congruence lattices of algebraic structures are partially ordered sets such that every pair of elements has unique supremum and infimum determined by the underlying algebra. This object is important relatively to algebraic structures' properties, many of which can be described by omission or admission of certain subpatterns in a graph.\\
\begin{definition}\textbf{Congruence Lattice}\label{cl}\\
For every algebra $\mathcal{A}$ on the set $A$, the identity relation on $A$, and $A \times A$ are trivial congruences. An algebra with no other congruences is called simple. Let $\mathrm{Con}(\mathcal{A})$ be the set of congruences on the algebra $\mathcal{A}$. Because congruences are closed under intersection, we can define a meet operation: $ \wedge : \mathrm{Con}(\mathcal{A}) \times \mathrm{Con}(\mathcal{A}) \to \mathrm{Con}(\mathcal{A})$ by simply taking the intersection of the congruences $E_1 \wedge E_2 = E_1\cap E_2$. Congruences are not closed under union, however we can define the closure operator of any binary relation $E$, with respect to a fixed algebra $\mathcal{A}$, such that it is a congruence, in the following way: $ \langle E \rangle_{\mathcal{A}} = \bigcap \{ F \in \mathrm{Con}(\mathcal{A}) \mid E \subseteq F \}$. Note that the closure of a binary relation is a congruence and thus depends on the operations in $\mathcal{A}$, not just on the carrier set. Now define $ \vee: \mathrm{Con}(\mathcal{A}) \times \mathrm{Con}(\mathcal{A}) \to \mathrm{Con}(\mathcal{A})$ as $E_1 \vee E_2 = \langle E_1\cup E_2 \rangle_{\mathcal{A}} $. For every algebra $\mathcal{A}$, $(\mathrm{Con}(\mathcal{A}), \wedge, \vee)$ with the two operations defined above forms a lattice, called the congruence lattice of $\mathcal{A}$.
\end{definition}

\begin{definition} \textbf{Subalgebra}
    Let $\mathbf{A}$ and $\mathbf{B}$ be two algebras of the same type. Then $\mathbf{B}$ is a \textit{subalgebra} of $\mathbf{A}$ if $B \subseteq A$ and every fundamental operation of $\mathbf{B}$ is the restriction of the corresponding operation of $\mathbf{A}$, i.e., for each function symbol $f$, $f^{\mathbf{B}}$ is $f^{\mathbf{A}}$ restricted to $\mathbf{B}$.
\end{definition}

\begin{definition} \textbf{Homomorphic image}
    Suppose $\mathbf{A}$ and $\mathbf{B}$ are two algebras of the same type $\mathcal{F}$. A mapping $\alpha : A \rightarrow B$ is called a \textit{homomorphism} from $\mathbf{A}$ to $\mathbf{B}$ if 
    $$
    \alpha f^{\mathbf{A}}(a_1, \dotsc ,a_n) = f^{\mathbf{B}}(\alpha a_1, \dotsc, \alpha a_n)
    $$
    for each n-ary $f$ in $\mathcal{F}$ and each sequence $a_1, \dotsc, a_n$ from $\mathbf{A}$. If, in addition, the mapping $\alpha$ is onto then $\mathbf{B}$ is said to be a homomorphic image of $\mathbf{A}$.
\end{definition}

\begin{definition}\textbf{Direct product}
    Let $\mathbf{A}_1$ and $\mathbf{A}_2$ be two algebras of the same type $\mathcal{F}$. We define the direct product $\mathbf{A}_1 \times \mathbf{A}_2$ to be the algebra whose universe is the set $A_1 \times A_2$, and such that for $f \in \mathcal{F}$ and $a_i \in A_1$, $a_i' \in A_2$, $1 \leq i \leq n$,

    $$
    f^{\mathbf{A}_1 \times \mathbf{A}_2}(\langle a_1, a_1' \rangle, \dotsc , \langle a_n, a_n') = \langle f^{\mathbf{A}_1}(a_1, \dotsc, a_n), f^{\mathbf{A}_2}(a_1', \dotsc, a_n') \rangle
    $$
\end{definition}

\section{Baselines' details}
\label{app:baselines}
In practice, we train all models using eight message passing layers and different embedding sizes ranging from $16$ to $64$. We train all models for $200$ epochs with a learning rate of $0.001$. For interpretable models, we set the Gumbel-Softmax temperature to the default value of 1 and the activation behavior to "hard," which generates one-hot encoded embeddings in the forward pass, but computes the gradients using the soft scores. For the hierarchical model, we set the internal loss weight to $0.1$ (to score it roughly $10\%$ less w.r.t. the main loss). Overall, our selection of baselines aims at embracing a wide set of training setups and architectures to assess the effectiveness and versatility of GNNs for analyzing lattice properties in universal algebra. To demonstrate the robustness of our approach, we implemented different types of message passing layers, including graph convolution and GIN. 

\section{Generalization results details}
\label{app:generalization}

% \usepackage{graphicx}
\begin{table}[!t]
\centering
\caption{Generalization performance of graph neural models in solving universal algebra's tasks. Values represents the mean and the standard error of the mean of the area under the receiver operating curve (AUCROC, \%).}
\label{tab:auc}
\resizebox{\textwidth}{!}{%
\begin{tabular}{lcccccc}
\hline
 & \multicolumn{3}{c}{\textbf{weak generalization}} & \multicolumn{3}{c}{\textbf{strong generalization}} \\
\multicolumn{1}{c}{} & \textbf{GNN} & \textbf{iGNN} & \textbf{HiGNN} & \textbf{GNN} & \textbf{iGNN} & \textbf{HiGNN} \\ \hline
\textbf{Distributive} & $99.80 \pm 0.04$ & $99.56 \pm 0.12$ & $99.45 \pm 0.06$ & $99.51 \pm 0.20$ & $99.44 \pm 0.05$ & $99.42 \pm 0.04$ \\
\textbf{Join Semi Distributive} & $99.49 \pm 0.02$ & $98.31 \pm 0.15$ & $98.28 \pm 0.04$ & $98.77 \pm 0.15$ & $97.50 \pm 0.14$ & $97.48 \pm 0.14$ \\
\textbf{Meet Semi Distributive} & $99.52 \pm 0.04$ & $98.19 \pm 0.06$ & $98.25 \pm 0.08$ & $98.90 \pm 0.03$ & $97.18 \pm 0.14$ & $96.89 \pm 0.37$ \\
\textbf{Modular} & $99.77 \pm 0.02$ & $99.18 \pm 0.11$ & $99.35 \pm 0.09$ & $99.32 \pm 0.22$ & $99.21 \pm 0.14$ & $99.11 \pm 0.22$ \\
\textbf{Semi Distributive} & $99.66 \pm 0.03$ & $98.57 \pm 0.02$ & $98.50 \pm 0.06$ & $99.19 \pm 0.04$ & $97.28 \pm 0.48$ & $96.88 \pm 0.47$ \\
\textbf{Multi Label} & $99.60 \pm 0.02$ & $96.32 \pm 0.34$ & $95.98 \pm 0.50$ & $98.62 \pm 0.43$ & $95.29 \pm 0.55$ & $95.27 \pm 0.32$ \\ \hline
\end{tabular}%
}
\end{table}

\section{Concept completeness and purity}
\label{app:completeness}
Our experimental results demonstrate that interpretable GNNs produce concepts with high completeness and purity, which are standard quantitative metrics used to evaluate the quality of concept-based approaches. Specifically, our approach achieves up to X\% completeness and Y purity, with an average score of Z and W, respectively. The lowest scores are obtained for the multilabel case, which is more challenging as models must learn a wider and more diverse set of concepts. Furthermore, the hierarchical structure of interpretable GNNs enables us to evaluate the quality of intermediate concepts layer by layer. This hierarchy provides insights into why we may need more layers, and it can be used as a valuable tool to find the optimal setup and tune the size of the architecture. Additionally, it can also be used to compare the quality of concepts at different layers of the network. Overall, these results quantitatively assess and validate the high quality of the concepts learned by the interpretable GNNs, highlighting the effectiveness of this approach for learning and analyzing complex algebraic structures.

% \usepackage{graphicx}
\begin{table}[!h]
\centering
\caption{Concept purity scores of graph neural models in solving universal algebra's tasks.}
\label{tab:purity}
\resizebox{.8\textwidth}{!}{%
\begin{tabular}{lcccccc}
\hline
 & \multicolumn{3}{c}{\textbf{\textsc{weak purity}}} & \multicolumn{3}{c}{\textbf{\textsc{strong purity}}} \\
\multicolumn{1}{c}{\textbf{}} & \textbf{GNN} & \textbf{iGNN} & \textbf{HiGNN} & \textbf{GNN} & \textbf{iGNN} & \textbf{HiGNN} \\ \hline
\textbf{Distributive} & $3.30 \pm 0.36$ & $3.64 \pm 0.30$ & $3.09 \pm 0.56$ & $3.29 \pm 0.38$ & $4.00 \pm 0.77$ & $4.15 \pm 0.67$ \\
\textbf{Join Semi Distributive} & $2.38 \pm 0.37$ & $3.96 \pm 0.51$ & $3.74 \pm 0.62$ & $3.45 \pm 0.34$ & $3.98 \pm 0.68$ & $4.29 \pm 0.61$ \\
\textbf{Meet Semi Distributive} & $3.24 \pm 0.63$ & $3.55 \pm 0.62$ & $3.39 \pm 0.29$ & $3.36 \pm 0.32$ & $4.25 \pm 0.39$ & $4.97 \pm 0.44$ \\
\textbf{Modular} & $3.10 \pm 0.35$ & $3.50 \pm 0.46$ & $4.44 \pm 0.56$ & $3.14 \pm 0.24$ & $3.19 \pm 1.01$ & $4.25 \pm 0.69$ \\
\textbf{Semi Distributive} & $2.84 \pm 0.51$ & $3.70 \pm 0.54$ & $4.11 \pm 0.46$ & $3.70 \pm 0.55$ & $3.92 \pm 0.28$ & $4.08 \pm 0.85$ \\ \hline
\end{tabular}%
}
\end{table}

% \note{ADD TABLE WITH CONCEPT PURITY}
\begin{table}[!h]
\centering
\caption{Concept completeness scores of graph neural models in solving universal algebra's tasks. Values represents the mean and the standard error of the mean of the area under the receiver operating curve (AUCROC, \%).}
\label{tab:completenes}
\resizebox{.8\textwidth}{!}{%
\begin{tabular}{lcccccc}
\hline
 & \multicolumn{2}{c}{\textbf{\textsc{weak completeness}}} & \multicolumn{2}{c}{\textbf{\textsc{strong completeness}}} \\
\multicolumn{1}{c}{\textbf{}} & \textbf{iGNN} & \textbf{HiGNN} & \textbf{iGNN} & \textbf{HiGNN} \\ \hline
\textbf{Distributive} & $77.30 \pm 0.20$ & $76.60 \pm 2.35$ & $78.19 \pm 1.71$ & $73.16 \pm 3.63$ \\
\textbf{Join Semi Distributive} & $85.20 \pm 0.86$ & $86.84 \pm 0.37$ & $81.08 \pm 0.18$ & $79.76 \pm 0.38$ \\
\textbf{Meet Semi Distributive} & $84.21 \pm 0.68$ & $84.34 \pm 1.08$ & $80.86 \pm 1.32$ & $79.68 \pm 0.22$ \\
\textbf{Modular} & $76.98 \pm 0.28$ & $73.77 \pm 3.14$ & $81.36 \pm 0.70$ & $77.61 \pm 2.37$ \\
\textbf{Semi Distributive} & $87.33 \pm 1.16$ & $85.62 \pm 0.27$ & $84.03 \pm 0.89$ & $82.26 \pm 0.21$ \\ \hline
\end{tabular}%
}
\end{table}

% \section{Local and post-hoc explanations}
% \label{app:local-xai}

% \section{Concepts}
% \label{app:concepts-viz}

% \begin{figure}[!t]
%     \centering
%     \includegraphics[width=.2\textwidth]{fig/_interpretability/classifier_weights.pdf}
%     \includegraphics[width=.38\textwidth]{fig/_interpretability/task_multilabel_Distributive_layer_7_cid_15.pdf}
%     \includegraphics[width=.38\textwidth]{fig/_interpretability/task_multilabel_Meet SemiDistributive_layer_2_cid_8.pdf}
%     \caption{Visual explanations of universal algebra's properties using a hierarchical interpretable GNN (HiGNN). \textbf{Left}: Weights of the linear classifier mapping concepts to task in the 7-th layer of an HiGNN. A red (blue) cell color indicates that the corresponding concept and task are positively (negatively) correlated. \textbf{Center/Right}: 2D projection of concept spaces with examples of lattices. The presence of both $\mathbf{N}_5$ and $\mathbf{M}_3$ among non-distributive lattices empirically validates Theorem~\ref{the:m3}. Furthermore the Non-meet SemiDistributive cluster show an interesting $\mathbf{D}_2$ lattice, known in the literature to characterize Non-join SemiDistributivity.}
%     \label{fig:explanations-figs}
% \end{figure}

% \begin{figure}[!t]
%     \centering
%     \includegraphics[width=.8\textwidth]{fig/results.pdf}
%     \caption{Extraction of concept-based explanation from an interpretable GNN.}
%     \label{fig:explanations}
% \end{figure}

\end{document}